\documentclass[10pt,twocolumn,letterpaper]{article}

\usepackage{cvpr}              

\usepackage{booktabs}
\usepackage{times}
\usepackage{amsmath,amssymb}
\usepackage{empheq}
\usepackage{graphicx}
\usepackage[font=normalsize]{caption}
\usepackage{subcaption}
\usepackage{floatrow}
\usepackage{enumitem}
\usepackage{xcolor}
\usepackage{multicol}
\usepackage{multirow}
\usepackage{tikz}
\usepackage{graphicx}
\usepackage{array}
\usepackage{colortbl}
\usepackage{tcolorbox}
\usepackage{makecell}
\usepackage{newtxtext,newtxmath}
\usepackage[accsupp]{axessibility}

\definecolor{cvprblue}{rgb}{0.21,0.49,0.74}
\usepackage[pagebackref,breaklinks,colorlinks,allcolors=cvprblue]{hyperref}

\def\paperID{2253} 
\def\confName{CVPR}
\def\confYear{2025}

\title{Generative Photography: Scene-Consistent Camera Control\\ for Realistic Text-to-Image Synthesis}

\author{Yu Yuan$^{1}$, Xijun Wang$^{1}$, Yichen Sheng$^2$, Prateek Chennuri$^1$, Xingguang Zhang$^1$, Stanley Chan$^1$ \\
$^1$School of ECE, Purdue University 
$^2$NVIDIA
}

\begin{document}

\twocolumn[{%
    \renewcommand\twocolumn[1][]{#1}%
    \maketitle
    \begin{center}
        \captionsetup{type=figure}
        \includegraphics[width=1\textwidth, trim={0 0 0 0}, clip]{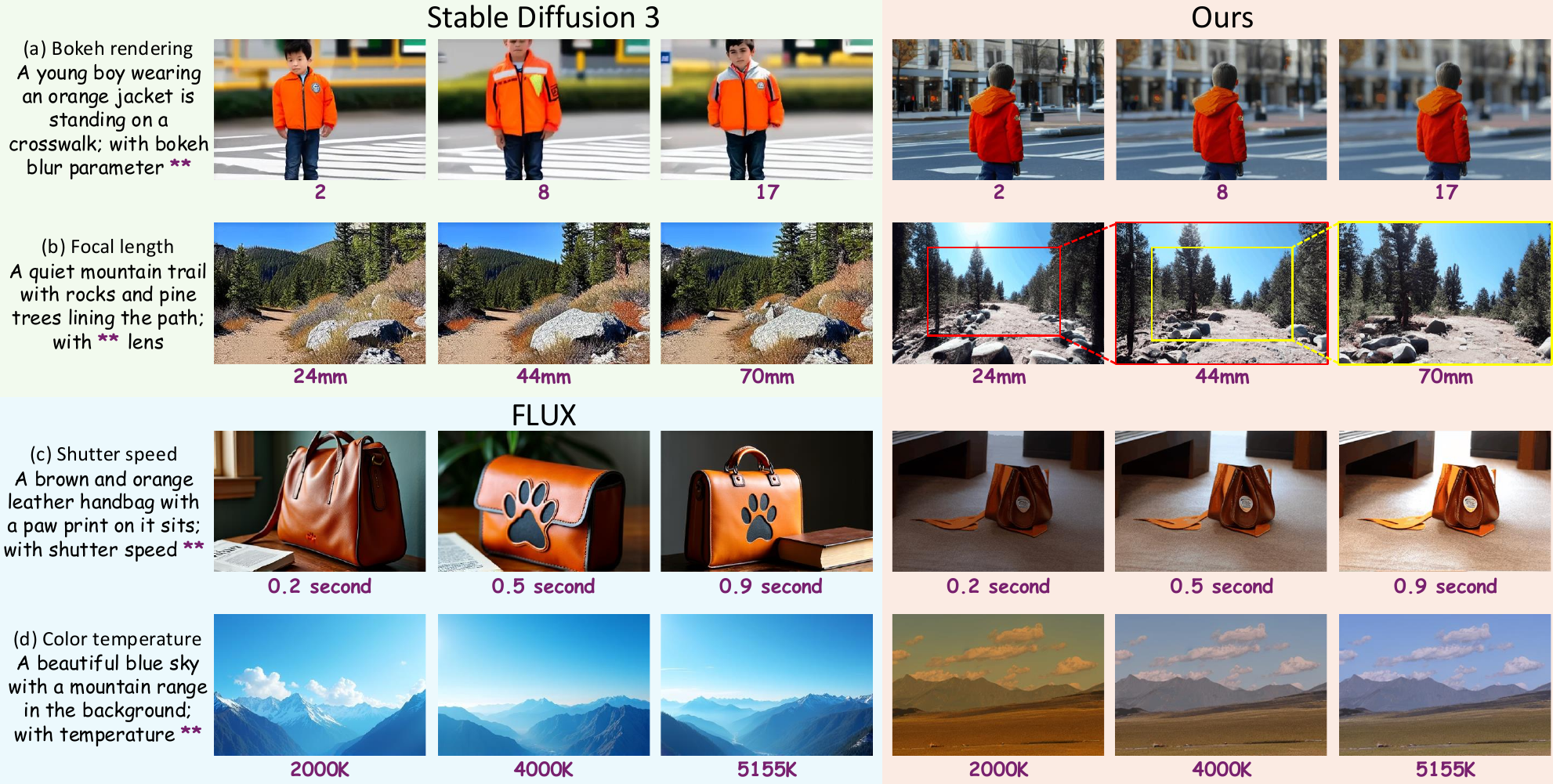}
        \caption{\small{This paper introduces generative photography, a new paradigm for text-to-image generation that maintains a consistent base scene while modifying only the camera settings to achieve varied photographic effects. Current state-of-the-art text-to-image generation models like Stable Diffusion 3 (SD3) \cite{SD2} and FLUX \cite{FLUX} face two major limitations: failure to accurately interpret camera-specific settings and difficulties in maintaining consistency in the base scene. This paper introduces a novel approach that addresses these issues, enabling precise camera setting control and maintaining scene consistency in generative models.}}
        \label{fig:teaser}
    \end{center}

}]

\begin{abstract}
Image generation today can produce somewhat realistic images from text prompts. However, if one asks the generator to synthesize a specific camera setting such as creating different fields of view using a 24mm lens versus a 70mm lens, the generator will not be able to interpret and generate scene-consistent images. This limitation not only hinders the adoption of generative tools in professional photography but also highlights the broader challenge of aligning data-driven models with real-world physical settings.
In this paper, we introduce \textbf{Generative Photography}, a framework that allows controlling camera intrinsic settings during content generation. The core innovation of this work are the concepts of Dimensionality Lifting and Differential Camera Intrinsics Learning, enabling smooth and consistent transitions across different camera settings. Experimental results show that our method produces significantly more scene-consistent photorealistic images than state-of-the-art models such as Stable Diffusion 3 and FLUX. Our code and additional results are available at \href{https://generative-photography.github.io/project}{https://generative-photography.github.io/project}.
\end{abstract}    
\section{Introduction}
\label{sec:intro}
Since the breakthrough of probabilistic diffusion models in the early 2020's \cite{Ho_2020_DDPM, Song_2021_Scorebased, Rombach_2022_LDM, Dhariwal_2021_Diffusion}, foundational vision models have created unprecedented opportunities for artificially generated content \cite{SORA, Blattmann_2023_SVD, Yang_2024_DAV, Guo_2023_AnimateDiff, Kondratyuk_2024_VideoPoet, Zhang_2023_Controlnet, Ruiz_2023_DB, Gal_2022_Textual, Rombach_2022_LDM, Ramesh_2021_DALLE, Saharia_2022_Imagen, Nichol_2022_Glide, SD2, FLUX}. Content generation processes that would have taken days of human labor in the past are now possible by machines in just a few seconds. 
Although current text-to-image models can produce perceptually realistic images, professional photographers remain skeptical, as these tools still cannot reliably replicate even basic camera effects.
For example, the prompt “\texttt{quiet mountain trail, 24mm lens}” and another prompt “\texttt{quiet mountain trail, 70mm lens}” will have no differences in the field of view (FoV) but different rocks and trees, as illustrated in Fig. \ref{fig:teaser} (b).

The inability of current diffusion models to accurately interpret camera settings reflects not merely limitations of scaling laws \cite{Kaplan_2020_Scaling}, but a fundamental gap between data-driven models and physical-world principles.
Recent research \cite{Kang_2024_Farvideogenerationworld} demonstrates that increasing training data and model parameters are insufficient for generation models to capture essential physics principles. To bridge this gap in photography context, we aim to \textit{teach camera physics} to the models. 
However, teaching camera physics is not a trivial task due to the lack of training data and the fundamental problem of existing image/video generation foundation models~\cite{Kang_2024_Farvideogenerationworld}.  
This paper introduces a novel framework, jointly designed through data curation and network architecture, to enable Generative Photography—specifically targeting image generation with camera intrinsics controls.

\textbf{Why do Existing Generative Models Fail to Understand Camera Physics?} There are two primary reasons why existing generative models fail: (1) The available \underline{training data is very limited}. To train a generative photography model, we need sets of images captured under different camera settings, such as varying apertures or focal lengths. Apart from the fact that acquiring these paired images is tedious and time/manpower-consuming, there is also a lack of textual labels. Therefore, in the absence of sufficient training data, it is challenging to train or fine-tune existing visual models; (2) Even if we have the training data, it remains unclear how one can \underline{disentangle the information} of scene embeddings from the camera embeddings. If we cannot disentangle these embeddings, then it will be impossible to keep the scene unchanged while altering the camera settings.
This would lead to severe scene consistency problems as illustrated in Fig.~\ref{fig:teaser}.

\textbf{What is Generative Photography?} Generative photography is an emerging approach in photography where content is generated instead of physically captured. 
Generative photography stresses \emph{camera-awareness}. On top of an existing text-to-image generation process, we demand the model to comprehend the typical camera settings: adjusting the aperture, shutter speed, focal length, and color temperature. By adjusting these camera settings, we can generate a variety of photorealistic effects such as different bokeh effects, exposure, color, and zoom, unlocking new applications in photography. A successful generative photography method should satisfy three objectives: (1) the camera effects are realistically rendered; (2) by changing the camera settings, the content of the scene is not altered, e.g., the buildings remain the same buildings and persons remain the same persons; (3) adding camera awareness does not degrade the image quality when compared to the baseline models that do not possess this property.

In this paper, we highlight the importance of maintaining scene consistency across varying camera settings. For example, transitioning from a "24mm lens" to a "70mm lens" should result in a smooth and consistent change in the field of view (FoV). This seamless transition between different camera intrinsics introduces the concept of a \textit{camera dimension}.  Analogous to spatial and temporal dimensions, generative photography is to solve consistency  problem in spatial and camera dimensions.    
To handle this camera dimension effectively, we propose two complementary approaches: \textbf{Dimensionality Lifting}, transforming text-to-image generation from a purely spatial problem into a joint space-camera domain for improved camera intrinsics disentanglement and spatial consistency; and \textbf{Differential Camera Intrinsics Learning}, a strategy designed to explicitly capture differences in camera intrinsic settings, operating simultaneously at both data and network architecture levels to reinforce consistent scene representation.

In summary, the contribution of this paper is two-fold:
\begin{enumerate}
\item We introduce the concept of Generative Photography, a new paradigm focusing on enabling text-to-image diffusion models with precise and consistent controllability over camera intrinsic settings.
\item To support generative photography, we present two new techniques, dimensionality lifting and differential camera intrinsics learning, to allow scene-consistent camera control for realistic text-to-image generation. This framework can be used on various camera effects, producing appealing camera control with consistent scenes.
\end{enumerate}

\section{Prior Work and Limitations}
\label{sec:related_works}
\subsection{Text-to-image Generation} 
Diffusion models \cite{Ho_2020_DDPM, Song_2021_Scorebased, Rombach_2022_LDM, Dhariwal_2021_Diffusion} have emerged as a powerful framework in the field of generative AI, primarily due to their robustness and versatility in generating high-quality images from textual descriptions. Tutorials such as \cite{Chan_2024_Tutorial, Kingma_2019_Tutorial, Luo_2022_Understanding} have provided comprehensive coverage of the foundations of these diffusion models. 

Notable implementations of text-to-image generation utilizing diffusion models include DALL-E \cite{Ramesh_2021_DALLE}, Stable Diffusion \cite{SD2}, DreamBooth \cite{Ruiz_2023_DB}, IMAGEN \cite{Saharia_2022_Imagen}, FLUX \cite{FLUX}, and others \cite{Rombach_2022_LDM, Gal_2022_Textual, Nichol_2022_Glide}. DALL-E pioneered the generation of novel images by combining concepts, attributes, and styles derived from textual input, demonstrating an advanced capability for understanding and creatively interpreting descriptions. Stable Diffusion further enhances this by generating high-resolution images in a latent space, significantly improving both computational efficiency and output quality.

\subsection{Camera-Awareness} 
When we think about camera-aware content generation, a fundamental question is how to control the diffusion model so that it respects the camera settings. The following three families of approaches are the most relevant ones to us:

\textbf{Camera-Awareness Via Guidance.} In theory, camera awareness can be done by integrating the control signals into the generation process through guidance. Methods such as ControlNet \cite{Zhang_2023_Controlnet}, T2I-Adapter \cite{Mou_2023_T2Iadapter}, and GLIGEN \cite{Li_2023_Gligen} belong to this category where they use depth maps, edge maps, semantic maps, and object-bounding boxes to guide the diffusion process. However, this guidance is mostly about the \emph{scene}, not the \emph{camera}. We are interested in the latter.

\textbf{Camera External Parameters.} Existing work related to our work mostly focuses on controlling the camera external parameters such as its pose and trajectory \cite{Wang_2024_MotionCtrl, He_2024_Cameractrl, Xu_2024_Camco, Hou_2024_Camtrol, Marmon_2024_Camvig, Bahmani_2024_VD3D, Courant_2024_ET, Jiang_2024_CCD, Kuang_2024_CVD, Cheong_2024_Boostingcameramotioncontrol, Xu_2024_Cavia}. Approaches like CameraCtrl \cite{He_2024_Cameractrl} and CamCo \cite{Xu_2024_Camco} incorporate these external parameters into pre-trained generative models by leveraging an additional camera encoder network. However, these methods are limited to single-camera trajectories, leading to significant inconsistencies in content and dynamics when generating multiple videos. Collaborative video diffusion (CVD) \cite{Kuang_2024_CVD} and Civia \cite{Xu_2024_Cavia} improve consistency by aligning features across multi-view video generation branches. 

\textbf{Camera Intrinsic Settings.} When it comes to intrinsic camera settings, there is very limited work except for some basic zoom-in/out effects \cite{Guo_2023_AnimateDiff, Blattmann_2023_SVD, RunwayGen3, Sun_2024_Dimensionxcreate3d4d, Fang_2024_Camera}. One of the main challenges is that most datasets used for training vision models often lack comprehensive camera settings. Even when some camera metadata is included, there is typically an absence of multi-setting metadata for the same scene \cite{Fang_2024_Camera}. This lack of parameter variation means that the available metadata cannot fully capture the true physical meaning of these settings. For instance, for a given scene, having only a single 50mm f/8 setting makes it difficult to infer how the depth of field (DoF) and field of view (FoV) would change under a 16mm f/4 setting for the same scene.

\subsection{Information Disentanglement} Beyond the previous related works, a major difficulty in camera-aware content generation is the user-computer interface. In today's text-to-image generation, a user needs to modify the prompt to control the image attributes. However, once the camera prompt is modified, the overall image structure can be severely distorted due to the sensitivity of outputs to the prompt-seed combination \cite{Kawar_2023_Imagic, Ruiz_2023_DB, Wu_2023_Uncovering, Gandikota_2023_Sliders}. This happens because we are not able to disentangle the camera embeddings and scene embeddings.

\textbf{GAN-based Disentanglement.} Information disentanglement can be done in many ways. In the pre-diffusion era, generative adversarial networks (GANs) \cite{Goodfellow_2014_GAN} have shown potential for highly disentangled control in their latent spaces, enabling precise manipulation of facial attributes without affecting others \cite{Karras_2019_Stylegan, Shen_2020_Interfacegan, Shen_2020_Interpreting, Shen_2021_Close, Tov_2021_Stylegan2}. For example, StyleGAN \cite{Karras_2019_Stylegan, Tov_2021_Stylegan2} allows detailed control over image properties through linear editing of its latent space, isolating specific attributes while preserving others.

\textbf{Diffusion-based Disentanglement.} There is increasing evidence that diffusion models can perform some degree of latent space information disentanglement. For example, Wu \textit{et al.} \cite{Wu_2023_Uncovering} did a case study about Stable Diffusion \cite{SD2} and consequently proposed to optimize the text embeddings during inference to maintain image coherence. There are new mechanisms to explicitly enforce information disentanglement, e.g., Wu \textit{et al.} \cite{Wu_2024_Contrastive} used contrastive guidance by sending two prompts to disentangle the information, and Rohit \textit{et al.} \cite{Gandikota_2023_Sliders} minimized interference by identifying a low-rank direction. The general problem of these approaches is that they are mostly tailored to facial attributes. When applying them to generic image content, they struggle with shapes and colors. Additionally, they primarily focus on optimizing the text embeddings during inference without considering any joint strategies to decouple data and models.

\begin{figure*}[h]
\centering
\includegraphics[width=0.9\linewidth, trim={0 0 0 0}, clip]{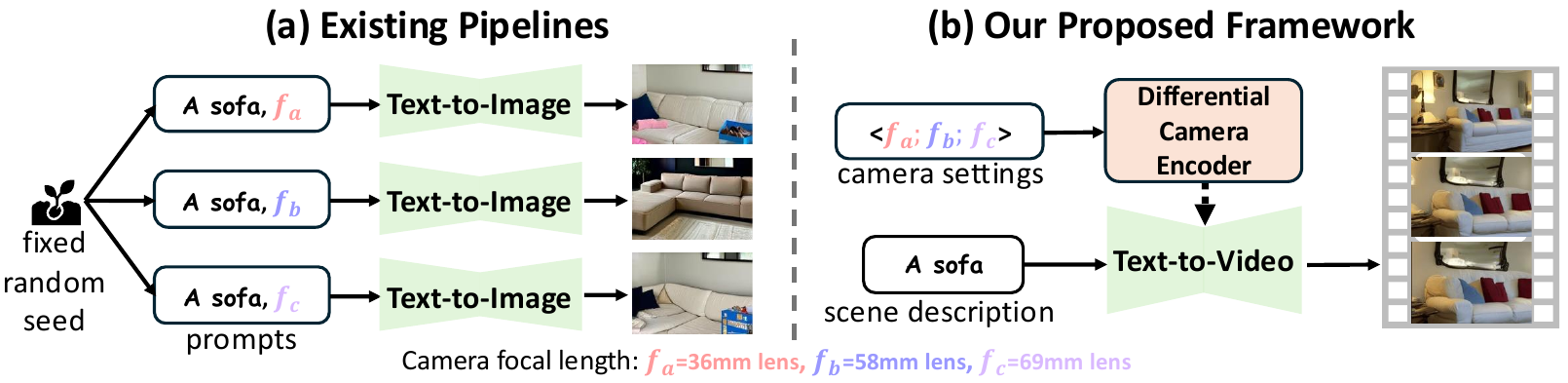}
\caption{\small{(a) Existing text-to-image (T2I) models struggle to perceive physical camera settings and maintain consistency across multiple settings, even when the random seed is fixed.
(b) We solve this problem by lifting camera-controlled text-to-image (T2I) generation into text-to-video (T2V) generation, thereby decoupling scene description from camera settings and achieving better scene consistency.}}
\label{fig: DL}
\end{figure*}
In this paper, we introduce the task of generative photography. We propose a framework consisting of two concepts: dimensionality lifting and differential camera intrinsics learning, which disentangle camera and scene embeddings. Our framework enables precise control over various camera settings, such as shutter speed, aperture, focal length, and color temperature, while maintaining scene consistency, allowing for exposure, bokeh rendering, zoom, and color control in generated images. These aspects will be detailed in the proposed methods section.
\section{Proposed Methods}
\label{sec:proposed_methods}
\subsection{Dimensionality Lifting}

A critical question in generative photography is how to disentangle the camera embedding from the scene embedding. Without this disentanglement, it becomes challenging to maintain consistency in the scene. In this paper, we first propose the idea of dimensionality lifting to achieve this. Figure \ref{fig: DL} (a) illustrates the low consistency between generated images in existing text-to-image (T2I) generation processes, where only part of the prompt is modified (in this case, the camera focal length). This paper addresses this issue within a higher-dimensional text-to-video (T2V) paradigm. As shown in Fig. \ref{fig: DL} (b), we lift multi-camera setting image generation with video generation. Within the T2V framework, we decouple the invariant scene description from camera settings: the scene description is used to establish a foundational scene, while the camera settings provide extra constraints for each corresponding frame.

There are several reasons why dimensionality lifting could solve our problem: (i) Variations in camera settings are important factors to consider; thus, we should elevate our base problem from a space-only problem to a space-camera joint problem.
(ii) This approach allows for a modular separation between scene elements and camera settings, enabling more flexible and accurate manipulations of camera-related aspects without impacting the underlying scene structure. 
(iii) Video generation models, due to their spatiotemporal attention design, are inherently better at maintaining consistency across frames \cite{Ho_2022_VDM, Blattmann_2023_SVD, Xie_2024_SV4D, Shi_2023_MVDream}.  (iv) We can leverage the powerful generative capabilities of pre-trained T2V models to ensure high-quality image generation.

\subsection{Differential Camera Intrinsics Learning}
Another challenge in generative photography is teaching the model to comprehend the physical settings of a camera. To address this, we introduce a differential camera intrinsics learning approach comprising two key components: differential data and a differential camera encoder. On the data side, we construct a differential dataset containing image-text pairs of a fixed scene captured under varying camera settings. At the network level, we incorporate the differences between adjacent camera settings to enhance the model's understanding of their effects.

\subsubsection{Differential Data}
\label{differential_data}
The differential dataset provides data-driven support for camera settings understanding in generative tasks.

\textbf{What is Differential Data and Why is it Useful?} Differential data involves creating pairs or sets of images that highlight specific differences, such as varying camera settings, while keeping other elements constant. Differential data is advantageous as it enables models to focus on and learn from the differences between images, rather than being overwhelmed by the vast diversity of real-world data. Here is a simple example of differential data.
\begin{center}
\begin{tcolorbox}[
  colframe=black, 
  colback=green!5, 
  boxrule=1pt, 
  sharp corners, 
  width=0.92\linewidth 
]
\textbf{Example.} If the images vary by color temperature, their corresponding textual labels are:

Invariant scene description: \texttt{"A squirrel eating a leaf from a tree."}
 \\
Color temperature for each frame: \textless \texttt{\textit{9933K}; \textit{3626K}; \textit{6302K}; \textit{4039K}; \textit{2400K}} \textgreater
\end{tcolorbox}
\end{center}

\begin{figure}[h]
\centering
\includegraphics[width=0.99\linewidth, trim={0 0 0 0}, clip]{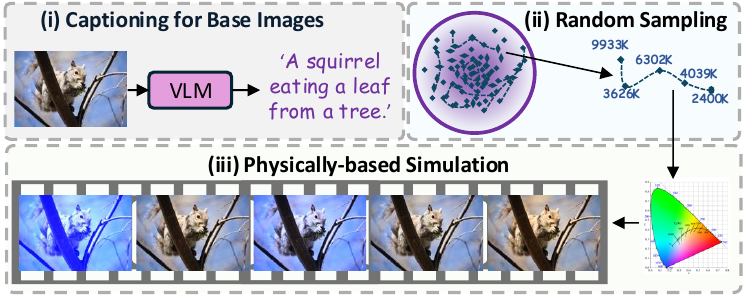}
\caption{\small{The pipeline of building differential data.}}
\label{fig: LLM}
\end{figure}

\textbf{How is the Differential Data Constructed?} We propose to construct a differential dataset in three steps following the outline shown in Fig. \ref{fig: LLM}.

\textbf{(i). Vision Language Model Captioning for Base Images:} To ensure that the data we generate can be used for training, we first need to collect base images and obtain their textual labels. The base images we select must be of high quality and meet specific requirements for camera generation tasks. For example, for shutter speed control, the base images should have a wide dynamic range and appropriate exposure; for color temperature control, all base images should lean towards a neutral color temperature; for bokeh rendering tasks, the base images should be all-in-focus images; and for focal length control, the base images need to have sufficiently high resolution. We employ a vision language model (VLM) LLaVA \cite{Liu_2023_Llava} to generate scene captions for a base image, which we use as the \underline{invariant scene descriptions} for this differential set.

\textbf{(ii). Random Sampling on Continuous Camera Space:} Camera settings are typically treated as continuous variables, such as focal length, shutter speed, and color temperature. We argue that each value on the camera setting scale is crucial, as it embodies a deeper physical understanding rather than merely representing a few discrete values. Continuous sampling supports continuous-scale training \cite{Chai_2022_Anyresolution, Chen_2024_Image, Wolski_2024_Scale}, which enhances the model's ability to perform across any value on the scale. Therefore, when constructing differential data, we perform random sampling across the continuous camera setting scale to gather multiple values. Specifically, as shown in Fig. \ref{fig: LLM}, we assume an \textit{invariant scene description} on a continuous camera setting scale (here exemplified by color temperature, with the vertical axis representing color temperature). For each instance, we sample several (5 here) points along the horizontal axis and use this collection of points as a set to describe camera setting variations.
We did not order the randomly collected data points in each set, ensuring that our network does not merely fit a specific directional pattern (such as simple zoom in/out for focal length) but instead learns the numerical values themselves. For the camera's color temperature, we collect a range from 2,000 to 10,000 Kelvin. For shutter speed, we re-normalize it to a scale of 0.1 to 1.0 (where larger values correspond to brighter images). For focal length, we sample values between 24 mm and 70 mm. Regarding bokeh rendering, we sample blur parameter values from 1 to 30, with larger values corresponding to stronger blur effects.

\textbf{(iii). Physically-based Simulation:} In the context of physically-based simulation, after obtaining the randomly sampled camera settings, we render the corresponding frames based on physical simulation for the base image. 
\begin{itemize}
    \item Bokeh rendering \cite{Sheng_2024_Bokeh, Peng_2022_BokehMe, Peng_2022_MPIB}: We first extract the base image's depth map using a depth estimation model Depth Anything \cite{Yang_2024_Depthanythingv1, Yang_2024_Depthanythingv2} and convert it to a disparity map, which, together with the base image and the sampled blur parameter, is input to the bokeh rendering module \cite{Peng_2022_BokehMe}. During this process, the refocused disparity remains fixed at the foreground depth, ensuring the foreground subject remains sharp while only adjusting the background blur.
    \item Focal length: We draw on Level-of-Detail methods \cite{Witkin_1987_Fliter, Mallat_1989_Wavelet}; specifically, for a given high-resolution base image, we calculate the field of view (FoV) ratio \cite{Ray_2002_Applied} corresponding to the sampled focal length relative to the original focal length (thus the sampled focal length must exceed the base focal length). Based on this ratio, we center-crop the corresponding region from the base image and subsequently resize all images captured at different focal lengths to the same resolution.
    \item Shutter speed: We obtain images at different exposure levels by adjusting the captured image irradiance and converting it back to RGB images using a simulated image signal processor (ISP) \cite{Li_2022_Efficient}.
    \item Color temperature: We adjust the RGB channel ratios according to the relationship between blackbody radiation and color temperature using empirical approximation \cite{Fairchild_2013_Color} to fit various color temperatures.
\end{itemize}
Through the aforementioned steps, we obtain multiple frames corresponding to variations in camera settings, along with the associated invariant scene description and camera settings. For further details, please refer to the appendix.

\subsubsection{Differential Camera Encoder} 

\begin{figure}
\centering
\includegraphics[width=0.99\linewidth, trim={0 0 0 0}, clip]{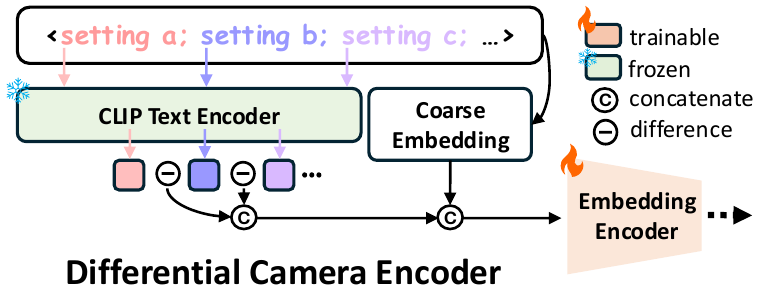}
\caption{\small{The overall architecture of differential camera encoder.}}
\label{fig: DCE}
\end{figure}

As shown in Fig \ref{fig: DL} (b), after dimensionality lifting, we use the scene description as the primary input to the T2V model to represent the base scene, with the camera settings as additional conditions. These settings are fed into the Differential Camera Encoder, which injects them into the foundation models to control the camera effects for each frame. In the differential camera encoder, we enhance the network's ability to understand and generate camera settings by embedding camera settings and providing additional differential information.

As illustrated in Fig. \ref{fig: DCE}, for each frame’s camera setting, we first create a coarse embedding, which incorporates a physical prior at the multi-channel pixel level (details in Appendix), similar to our physically-based simulation approach. Additionally, we capture inter-frame differences among camera settings as an auxiliary input to introduce finer-grained semantic difference information. Specifically, we utilize a frozen CLIP text encoder \cite{Radford_2021_CLIP} to generate camera features for each frame and compute the feature differences between adjacent values of camera settings. 

We then concatenate the coarse embedding with these differential features and feed them into an embedding encoder specifically designed for videos, similar to the T2I-Adaptor \cite{Mou_2023_T2Iadapter}. The embedding encoder and the T2V foundation encoder are structurally similar, which facilitates the integration of the multi-level features obtained from the embedding encoder into the foundation models.
Following recommendations from CameraCtrl \cite{He_2024_Cameractrl}, we avoid using ControlNet \cite{Zhang_2023_Controlnet} to mitigate information leakage issues.

We propose that the Differential Camera Encoder enhances the model’s ability to differentiate camera settings by focusing on comparative effects between them. This approach moves beyond a solely data-driven paradigm, significantly reducing the training data requirements and computational costs, while providing the network with more robust and precise control capabilities.
\section{Experiments}
In this section, we evaluate the application of our proposed framework in generative photography. Section \ref{implementation} presents implementation details, Section \ref{comparisons} compares our framework with other potential methods, and Section \ref{ablation} discusses the results of ablation studies.

\subsection{Implementation Details}
\label{implementation}
\textbf{Datasets.} As mentioned in Sub-section \ref{differential_data}, we need high-quality base images with task-related features to simulate camera effects. For the exposure and white balance conditioning tasks, our base images are sourced from our own photography as well as public image datasets \cite{Zhang_2019_Zoom, Ouyang_2021_Neural}. For the bokeh rendering task, we collected all-in-focus base images from our own photography and \cite{Hasinoff_2016_HDR, Zhang_2019_Dof, Ershov_2020_Problems}, ensuring they contained a well-defined foreground subject with a distinguishable background. For the focal length task, our base images were sourced from our own photography and \cite{Zhang_2019_Zoom, Ershov_2020_Problems}, with all images captured using a full-frame camera equipped with a 24mm lens. The base images for focal task have a minimum resolution of 3,000 on the shorter side to ensure high quality and detailed high-frequency information, even after partial central cropping. 

Each task includes 1,000 images with diverse scenes and varied shooting conditions. Note that these datasets are scalable, though we found 1,000 images sufficient to achieve the desired camera control effects.

\textbf{Training and Inference.} In this work, we select AnimateDiff \cite{Guo_2023_AnimateDiff} as our T2V base model, reducing the generated video frames to 5 to decrease computational costs for both training and inference. During training, most parameters of AnimateDiff are kept frozen, with only the motion LoRA \cite{Hu_2021_Lora} and our differential camera encoder being fine-tuned/trained. We use a learning rate of 1e-4 with the Adam optimizer. The video resolution for training is set to 256 × 384. For each task, we train with a batch size of 8 for 25,000 epochs, taking approximately 10 hours on an Nvidia A100 80GB GPU.

During inference, we input a prompt along with a set of camera settings to generate multiple frames. It’s worth noting that inference can also be conducted with a single camera setting by replicating this value across multiple frames, which results in a static output across frames. For bokeh rendering inference, no depth map is required.

\textbf{Metrics.} Our proposed evaluation metrics primarily focus on accuracy to physical laws and scene consistency. The accuracy of generated images with respect to camera physical settings is evaluated using the Pearson correlation coefficient (CorrCoef) of trend changes between the generated images and the reference video. For scene consistency, we use frame-wise Learned Perceptual Image Patch Similarity (LPIPS) \cite{Zhang_2018_Perceptual} to calculate the perceptual feature distance between frames generated with different camera settings. Additionally, to evaluate the impact of added camera controls on generation following, we use CLIP \cite{Radford_2021_CLIP} to compute the similarity score between generated images and the text prompt. For more details on the metrics, please refer to the supplementary material. 

For each task, we chose 75 sets of camera settings for testing. These sets emphasize a wide range settings to enhance the robustness and persuasiveness of the evaluation.

\subsection{Comparisons with Other Methods}
\label{comparisons}

\begin{figure*}[h]
\centering
\includegraphics[width=\linewidth]{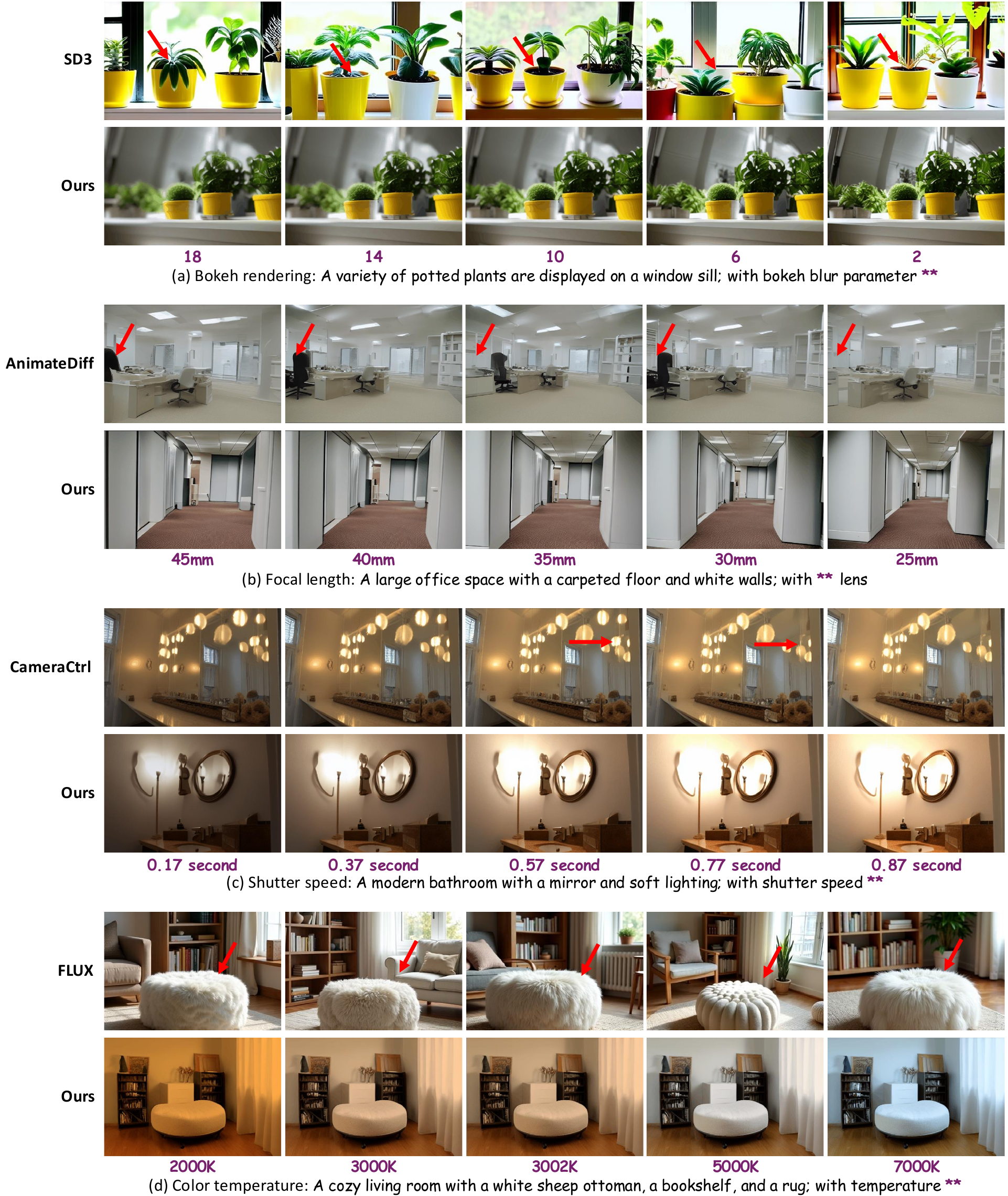}
\caption{\small{Visual comparisons between different generative methods. Our method is capable of generating realistic camera effects for any given camera setting scale, while maintaining high scene consistency across images corresponding to different scales. Both AnimateDiff \cite{Guo_2023_AnimateDiff} and CameraCtrl \cite{He_2024_Cameractrl} have been fine-tuned on our data. We highlight the discontinuities in the scene with red arrows.}}
\label{fig: baselines}
\end{figure*}

\begin{table*}[ht]
    \centering\resizebox{\textwidth}{!}{
    \small
    \begin{tabular}{l|ccc|ccc|ccc|ccc}
        \Xhline{3\arrayrulewidth}
      \multirow{3}{*}{Methods} 
           & \multicolumn{3}{c|}{Bokeh Rendering} & \multicolumn{3}{c|}{Focal Length} & \multicolumn{3}{c|}{Shuttle Speed} & \multicolumn{3}{c}{Color Temperature} \\
        \cline{2-13}
        & Accuracy & Consistency & Following
        & Accuracy & Consistency & Following
        & Accuracy & Consistency & Following
        & Accuracy & Consistency & Following\\
            
        & CorrCoef $\uparrow$ & LPIPS & CLIP $\uparrow$
        & CorrCoef $\uparrow$ & LPIPS  & CLIP $\uparrow$
        & CorrCoef $\uparrow$ & LPIPS  & CLIP $\uparrow$
        & CorrCoef $\uparrow$ & LPIPS  & CLIP $\uparrow$\\

        \hline
        \textit{Reference} & 1.0000 & 0.0527 &  0.3974 & 1.0000 & 0.4709 & 0.3853 &  1.0000 & 0.0511 & 0.3783 & 1.0000 & 0.0398 & 0.4053  \\
        
        \hline
        SD3 \cite{SD2} & 0.2492 & 0.7253 & ---- & 0.2356 & 0.7108 & ---- & 0.2731 & 0.6937 & ---- & 0.2312 & 0.6891 & ---- \\

        FLUX \cite{FLUX} & 0.2006 & 0.6770 & ---- & 0.2003 & 0.6461 & ---- & 0.2398 & 0.6403 & ---- & 0.2363 & 0.6155 & ---- \\

       AnimateDiff  \cite{Guo_2023_AnimateDiff} (w/o \textit{FT}) & 0.2960 & 0.1005  &  0.2753 & 0.2613 & 0.1208 & 0.2532 & 0.1843 & 0.1002 & 0.2631 & 0.1834 & 0.0805 & 0.2659 \\
        
       AnimateDiff \cite{Guo_2023_AnimateDiff} (w/ \textit{FT}) 
 & 0.3714 & \colorbox{blue!20}{0.0255}  &  0.2984 & 0.2597 & 0.2288 & 0.2739 & 0.2198 & 0.0948 & \colorbox{blue!20}{0.2936} & 0.2897 & \colorbox{blue!20}{0.0205} & 0.2839 \\
     
       CameraCtrl \cite{He_2024_Cameractrl} (w/o \textit{FT})  & 0.3303 & 0.1447 & 0.2804 & 0.2913 & 0.1144 & 0.2644 & 0.1896 & 0.0986 & 0.2912 & 0.1773 & 0.0935 & 0.2753 \\

       CameraCtrl \cite{He_2024_Cameractrl} (w/ \textit{FT})  & \colorbox{blue!20}{0.6025} & 0.1158 & \colorbox{red!20}{0.3017} & \colorbox{blue!20}{0.8671} & \colorbox{blue!20}{0.4606} & \colorbox{blue!20}{0.2865} & \colorbox{blue!20}{0.7526} & \colorbox{blue!20}{0.0775} & 0.2981 & \colorbox{blue!20}{0.5812} & 0.0651 & \colorbox{blue!20}{0.2885} \\
        
       Ours & \colorbox{red!20}{0.8626}  & \colorbox{red!20}{0.0788} & \colorbox{blue!20}{0.3007} & \colorbox{red!20}{0.9695} & \colorbox{red!20}{0.4647} & \colorbox{red!20}{0.2871} & \colorbox{red!20}{0.9264} & \colorbox{red!20}{0.0695} & \colorbox{red!20}{0.3015} & \colorbox{red!20}{0.8970} & \colorbox{red!20}{0.0499} & \colorbox{red!20}{0.2910} \\
        \Xhline{3\arrayrulewidth}
    \end{tabular}}
    \caption{Quantitative comparison with different generative methods. Accuracy is computed by comparing with reference videos; Consistency by frame-wise LPIPS; Prompt following by CLIP. \textit{Reference} refers to the results obtained from physical simulations. \textit{FT} denotes fine-tuning on our differential data. We highlight the \colorbox{red!20}{best} and \colorbox{blue!20}{second-best} values for each metric.}
    \label{tab:baselines}
\end{table*}

\begin{table*}[ht]
    \centering\resizebox{\textwidth}{!}{
    \small
    \begin{tabular}{l|ccc|ccc|ccc|ccc}
        \Xhline{3\arrayrulewidth}
      \multirow{3}{*}{Methods} 
           & \multicolumn{3}{c|}{Bokeh Rendering} & \multicolumn{3}{c|}{Focal Length} & \multicolumn{3}{c|}{Shuttle Speed} & \multicolumn{3}{c}{Color Temperature} \\
        \cline{2-13}

        & Accuracy & Consistency & Following
        & Accuracy & Consistency & Following
        & Accuracy & Consistency & Following
        & Accuracy & Consistency & Following\\
            
        & CorrCoef $\uparrow$ & LPIPS  & CLIP $\uparrow$
        & CorrCoef $\uparrow$ & LPIPS  & CLIP $\uparrow$
        & CorrCoef $\uparrow$ & LPIPS & CLIP $\uparrow$
        & CorrCoef $\uparrow$ & LPIPS  & CLIP $\uparrow$\\
        
       \hline
        \textit{Reference} & 1.0000 & 0.0527 &  0.3974 & 1.0000 & 0.4709 & 0.3853 &  1.0000 & 0.0511 & 0.3783 & 1.0000 & 0.0398 & 0.4053  \\
        
        \hline
        w/o differential & 0.7631 & 0.1239 & 0.2964 & 0.8594 & 0.4943 & \colorbox{blue!20}{0.2885} & 0.8864 & 0.0698 & 0.3007 & 0.8586 & 0.0699 & 0.2904 \\
        
       Discrete & 0.6774 & 0.0854 & 0.2948 &  0.7254 & 0.4173 & 0.2833 & 0.8384 & \colorbox{blue!20}{0.0584} & 0.2976 & 0.8140 & \colorbox{blue!20}{0.0524} & 0.2946 \\
        
       Ours (3 frames)  & 0.8355 & 0.0814 & 0.2895 & 0.9273 & 0.3921 & 0.2857 & 0.9073 & 0.0685 & 0.2873 & 0.8848 & 0.0835 & 0.2955 \\

       Ours (5 frames) & \colorbox{blue!20}{0.8626}  & \colorbox{blue!20}{0.0788} & \colorbox{red!20}{0.3007} & \colorbox{blue!20}{0.9695} & \colorbox{red!20}{0.4647} & 0.2871 & \colorbox{blue!20}{0.9264} & 0.0695 & \colorbox{blue!20}{0.3015} & \colorbox{blue!20}{0.8970} & \colorbox{red!20}{0.0499} & \colorbox{blue!20}{0.2980} \\
        
       Ours (7 frames) & \colorbox{red!20}{0.8835} & \colorbox{red!20}{0.0747} & \colorbox{blue!20}{0.2988} & \colorbox{red!20}{0.9783} & \colorbox{blue!20}{0.4813} & \colorbox{red!20}{0.2909} & \colorbox{red!20}{0.9294} & \colorbox{red!20}{0.0535} & \colorbox{red!20}{0.3034} & \colorbox{red!20}{0.9095} & 0.0581 & \colorbox{red!20}{0.2985} \\

        \Xhline{3\arrayrulewidth}
    \end{tabular}}
    \caption{Quantitative results of ablation study.}
    \label{tab:ablation}
\end{table*}

To evaluate our method, we conduct comparisons against three categories of approaches: first, the state-of-the-art text-to-image generation models, Stable Diffusion 3 (SD3) \cite{SD2} and FLUX \cite{FLUX}; second, the text-to-video generation model, AnimateDiff \cite{Guo_2023_AnimateDiff}, fine-tuned on our differential dataset to assess the performance of a purely data-driven approach; and finally, CameraCtrl \cite{He_2024_Cameractrl}, where we adapt the model to control intrinsic camera settings instead of external ones, and train it on our dataset. Together, these baseline models form a comprehensive evaluation framework, encompassing image and video generation, external camera parameter control, and other related aspects. 

Fig. \ref{fig: baselines} demonstrates that our method preserves excellent scene consistency while realistically simulating camera effects for any given setting. Notably, in bokeh rendering, although no depth information was provided during inference, the rendered scenes appear to exhibit depth awareness, consistently keeping the foreground sharp while only the background varies according to the bokeh blur parameter. In color temperature control, our model also shows fine-grained control capabilities, producing images at 3,000K and 3,002K with only minimal differences in color temperature.
From Table \ref{tab:baselines}, although SD3 and FLUX demonstrate a clear advantage in prompt following, it exhibits lower consistency and accuracy between frames with different camera settings. After fine-tuning using our comparative dataset, both AnimateDiff and CameraCtrl showed obvious improvements in accuracy. Our method shows a significant advantage in both the accuracy and consistency of generated camera settings. It also performs well in prompt following, suggesting that our approach maintains the model's overall generative capabilities without significant compromise. Here, a lower LPIPS score is not always better, as variations in color temperature, exposure, focal length, and bokeh can affect frame differences. The key metric is proximity to the reference videos. Compared to other base models, our approach still excels in prompt following, indicating that the additional camera control does not compromise generative performance.

\subsection{Ablation study}
\label{ablation}
\textbf{Differential Camera Encoder.} To assess the impact of differential network design, we removed inter-frame differences in the differential camera encoder. Table \ref{tab:ablation} (w/o differential) suggests that incorporating different information enhances the model's accuracy and consistency.

\textbf{Data Sampling Strategy.} To evaluate the impact of continuous data sampling in constructing the dataset, we compared it with the discrete (uniform) sampling method using 100 data points. Table \ref{tab:ablation} (Discrete) demonstrates that continuous sampling of data points better facilitates the network's understanding of these physical values.

\textbf{Number of Frames.} Our model can perform training and inference across different frame counts. To assess the potential impact of varying training frames, we trained our model using 3, 5, and 7 frames. As shown in Table \ref{tab:ablation}, increasing the frame number tends to improve the model’s accuracy and consistency. This also highlights a trade-off: while a higher number of camera settings offers users more options and improved performance, it also increases computational cost and processing time.

\begin{figure}[h]
\centering
\includegraphics[width=1\linewidth]{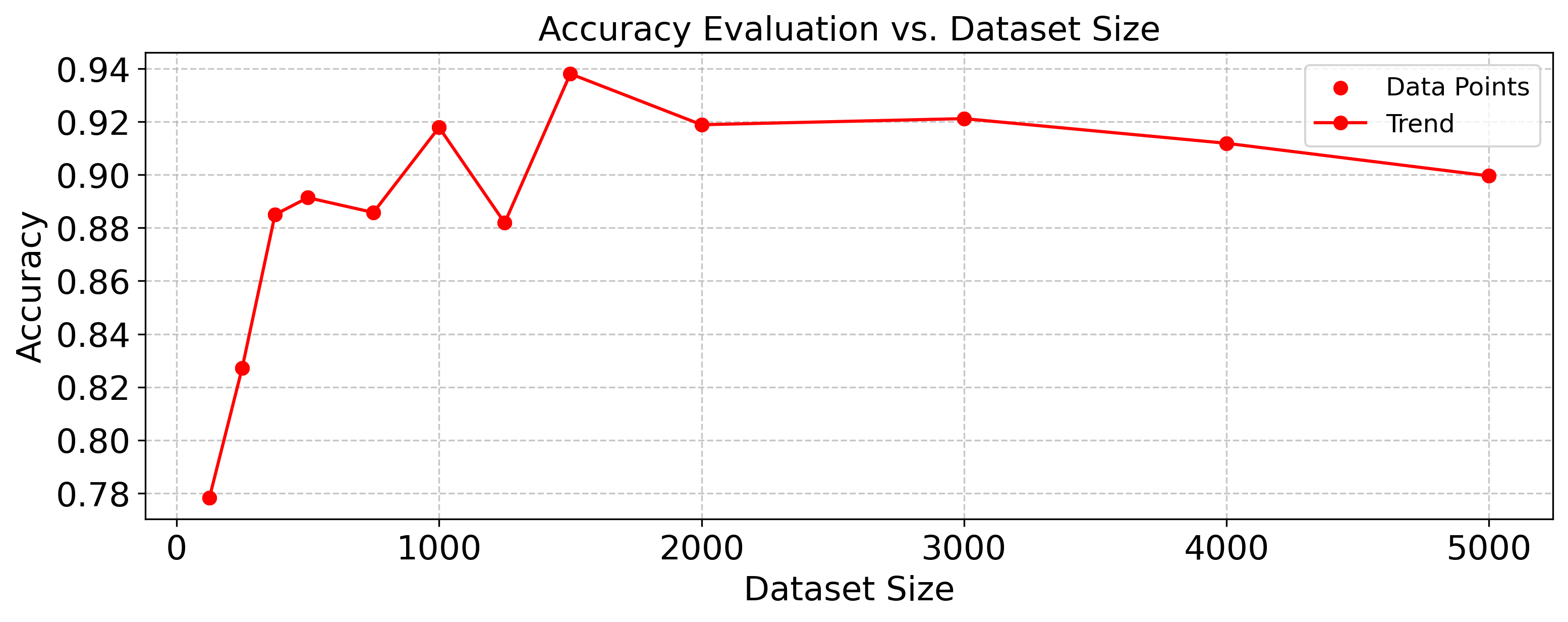}
\caption{\small{Ablation study on dataset scaling characteristic.}}
\label{fig: scale}
\end{figure}

\textbf{Dataset Scaling.} In Fig. \ref{fig: scale}, we present training dataset scaling experiments for shutter speed control. The results indicate that performance improves as the training data increases; however, beyond 1000 samples, the gains plateau. This suggests that our carefully designed dataset, combined with joint network optimization, is sufficient for understanding physical parameters and efficiently capturing fundamental principles with relatively little data.
\section{Conclusions}
Generative photography opens the door to a new paradigm in photography where the generated contents are consistent with the intrinsic camera physics. Our solution is based on two concepts: dimensionality lifting and differential camera intrinsics learning. It generates physically realistic camera effects while maintaining a high degree of scene consistency. Generative photography has the potential to reduce the post-processing burden in human-operated photography fundamentally. In addition, it offers a fresh perspective for generative models to understand the world better.

\textbf{Acknowledgements.} This work is supported, in part, by the National Science Foundation under grants 2133032, 2134209, 2030570, and 2431505.

\newpage

{
    \small
    \bibliographystyle{ieeenat_fullname}
    \bibliography{main}
}

\clearpage
\setcounter{page}{1}
\maketitlesupplementary

\section{Introduction}

This supplementary material provides additional discussions and details on the construction of differential data (Section \ref{sec:data}), network design (Section \ref{sec:network}), evaluation metrics (Section \ref{sec:metrics}), and more visual results (Section \ref{sec:results}).  

To better illustrate the continuity and effects of camera intrinsic setting control, we highly recommend readers view the \textbf{Videos/GIFs} provided in the project page: \href{https://generative-photography.github.io/project/}{https://generative-photography.github.io/project/}.

\section{More Details of Building Differential Data}
\label{sec:data}
Our differential data pipeline dynamically generates training data by storing only base images and scene descriptions. Camera settings are sampled during training and simulated \textbf{on-the-fly} using physical principles, producing differential multi-frame data without pre-storing large video files. This also ensures continuous sampling of training data.

We provide below additional key considerations for constructing differential datasets for each type of camera setting, along with sample demonstrations.

\subsection{Differential Data for Bokeh Rendering}

As shown in Fig. \ref{fig: bokeh_simu}, to enhance the prominence of the bokeh rendering effect, we impose the following two requirements on the base images:  1). The images should be nearly all-in-focus.  2). They should exhibit significant depth differences, allowing clear distinction between foreground and background.  

We employ bokehMe \cite{Peng_2022_BokehMe} for realistic bokeh simulation. During this process, the value of the refocused disparity is consistently maintained at the depth of the foreground.

\setcounter{figure}{6}
\begin{figure}[h]
\centering
\includegraphics[width=0.99\linewidth, trim={0 0 0 0}, clip]{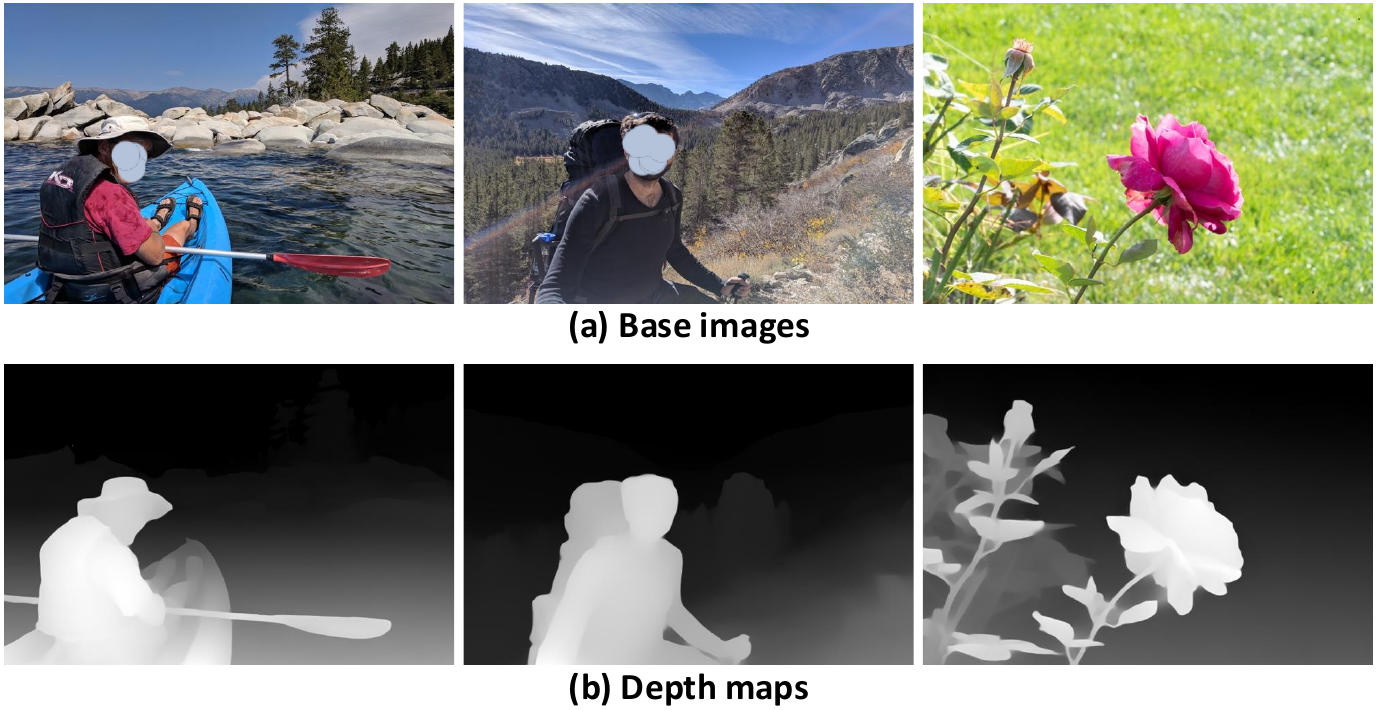}
\caption{\small{The first row shows examples of base images used for constructing bokeh rendering data, featuring prominent foregrounds and distinguishable backgrounds. The second row presents depth maps extracted using the Depth Anything \cite{Yang_2024_Depthanythingv1, Yang_2024_Depthanythingv2} model.}}
\label{fig: bokeh_simu}
\end{figure}

\subsection{Differential Data for Focal Length}

In the real world, obtaining a set of images of the same scene at multiple focal lengths is highly cumbersome, with a lack of perfect alignment between the images, and the achievable focal length range is limited \cite{Zhang_2019_Zoom}. In this paper, we reference the level-of-detail \cite{Witkin_1987_Fliter,  Mallat_1989_Wavelet} approach and compute the field-of-view (FoV) ratio of the desired focal length relative to the base image focal length. This ratio is then used for center cropping to approximate the actual continuous optical zoom process. In this subsection, we compare the performance of our method with that of actual optical zoom.

A camera's field-of-view (FoV) can be expressed in terms of the focal length \(f\) and the sensor dimensions (typically width \(w\) or height \(h\)). The formulas are as follows:

Horizontal FoV:  
\begin{equation}
\text{FoV}_\text{h} = 2 \cdot \arctan\left(\frac{w}{2f}\right)
\label{eq:fov_horizontal}
\end{equation}

Vertical FoV:  
\begin{equation}
\text{FoV}_\text{v} = 2 \cdot \arctan\left(\frac{h}{2f}\right)
\label{eq:fov_vertical}
\end{equation}

Diagonal FoV:  
\begin{equation}
\text{FoV}_\text{d} = 2 \cdot \arctan\left(\frac{\sqrt{w^2 + h^2}}{2f}\right)
\label{eq:fov_diagonal}
\end{equation}

where \(w\) denotes the width of the sensor, \(h\) is the height of the sensor, and \(f\) represents the focal length.

\begin{figure*}[h]
\centering
\includegraphics[width=1\linewidth, trim={0 0 0 0}, clip]{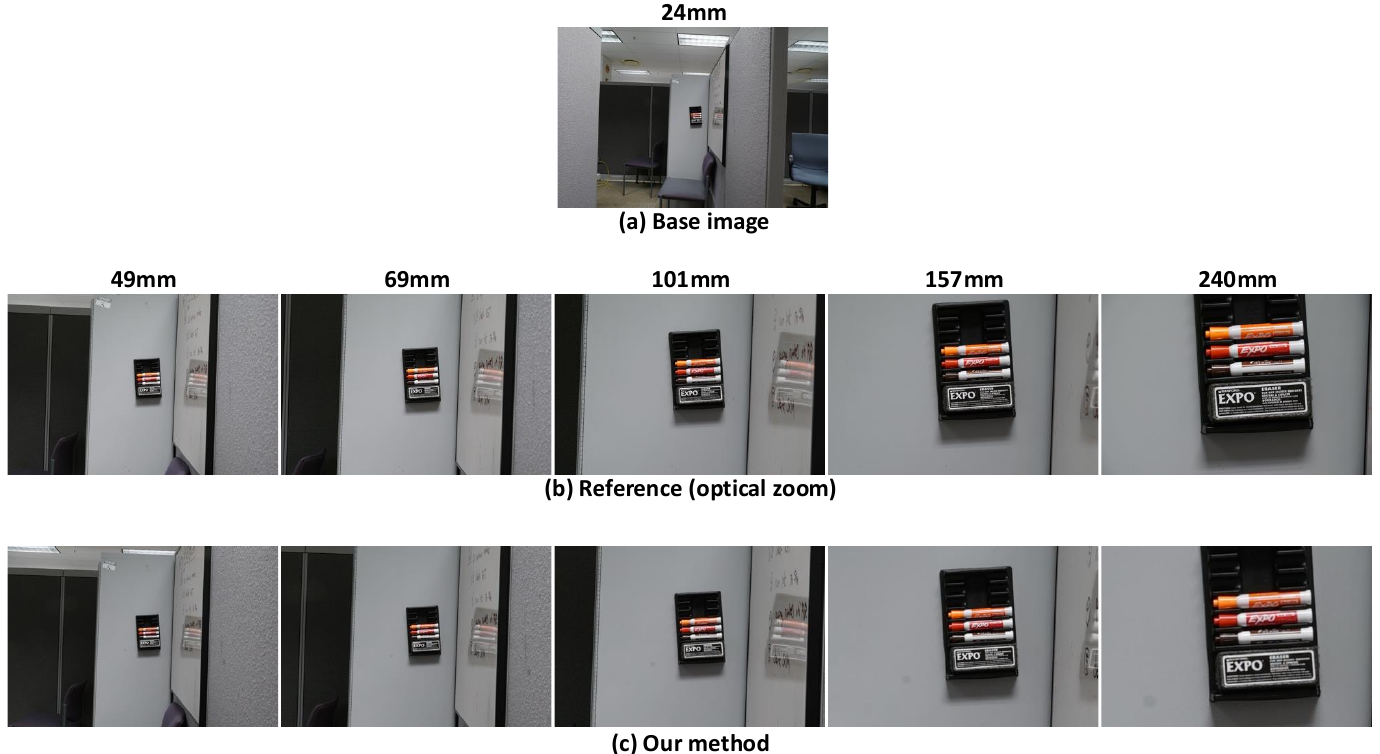}
\caption{\small{The comparison between the reference real focal lengths and our simulated results. Note that the real-world shooting data is derived from \cite{Zhang_2019_Zoom}, and there may be slight misalignment between images of different resolutions due to shooting conditions. We observe that excessively high focal length simulation ratios can lead to a decline in image quality. Therefore, in this study, the focal length range is constrained to 24-70mm. Please zoom in for a more detailed comparison.}}
\label{fig:zoom_simu}
\end{figure*}

Based on the aforementioned FoV calculation formula, we crop the central region of a high-resolution base image to simulate the corresponding view at larger focal lengths. Fig. \ref{fig:zoom_simu} compares the optical zoom with the results generated by our cropping method. The real-world data for different focal lengths is from \cite{Zhang_2019_Zoom}. Our method demonstrates a high degree of consistency with the real data in terms of FoV. It is worth noting that due to the resolution and quality constraints of the base image, excessive cropping leads to significant loss of detail and quality. Therefore, in this work, we limit the focal length range to 24-70mm.

\subsection{Differential Data for Shutter Speed}

A realistic imaging model can be formulated as follows, similar to \cite{Chi_2023_HDR, Qu_2024_Exposure, Li_2022_Efficient}. Consider a final LDR image, $L$, captured at an exposure time of $t$ where the underlying HDR scene irradiance map is represented by $H$. 
\begin{align}
L = \texttt{ADC}\bigg\{ & \xi \times \texttt{Clip}\big\{\text{Poisson}\big(t \times \text{QE} \times (H + \mu_{\text{dark}})\big)\big\} \nonumber \\
& + N(0, \sigma_{\text{read}}^2) \bigg\}^{1/\gamma}
\label{eq:imaging_model}
\end{align}

 where $\xi$ is the conversion gain, QE is the quantum efficiency, $\mu_{\text{dark}}$ is the dark current, and $\sigma_{\text{read}}$ is the read noise standard deviation. Here, $\text{Poisson}$ represents the Poisson distribution characterizing the photon arriving process and the dark current effect, and $N$ represents the Gaussian distribution characterizing the sensor noise. $\texttt{ADC}\left\{\cdot\right\}$ is the analog-to-digital conversion and $\texttt{Clip}\left\{\cdot\right\}$ is the full well capacity induced saturation effect. We assume a linear camera response function for CMOS sensors and that the imperfections in the pixel array, ADC, and color filter array have been mitigated.

For the shutter speed control task, we selected base images with a high dynamic range and appropriate exposure to approximate \( H \). By varying the parameter \( t \) in the formula \ref{eq:imaging_model}, we simulate multiple frames corresponding to different shutter speeds.

\subsection{Differential Data for Color Temperature}
We employ an empirical approximation revised from \cite{Fairchild_2013_Color} to map a given color temperature in Kelvin to corresponding RGB values, ensuring accurate and balanced color representation.  The input kelvin is normalized by dividing by 100, resulting in temp. The conversion process is as follows:

For \( \text{temp} \leq 66 \):
\begin{equation}
\begin{aligned}
    \mathbf{RGB} &= \left( 255, \right. \\
    &\quad \max\left(0, 99.47 \cdot \ln(\text{temp}) - 161.12\right), \\
    &\quad \left. \max\left(0, 138.52 \cdot \ln(\text{temp} - 10) - 305.04\right) \right)
\end{aligned}
\label{eq:awb_a}
\end{equation}

For \( 66 < \text{temp} \leq 88 \):
\begin{equation}
\begin{aligned}
    \mathbf{RGB} &= \left( 0.5 \cdot \left( 255 + 329.70 \cdot (\text{temp} - 60)^{-0.1933} \right), \right. \\
    &\quad  0.5 \cdot \left( 288.12 \cdot (\text{temp} - 60)^{-0.1155} \right. \\
    &\quad + 99.47 \cdot \ln(\text{temp}) - 161.12 \left. \right), \\
    &\quad \left. 0.5 \cdot \left( 138.52 \cdot \ln(\text{temp} - 10) - 305.04 + 255 \right) \right)
\end{aligned}
\label{eq:awb_b}
\end{equation}

For \( \text{temp} > 88 \):
\begin{equation}
\begin{aligned}
    \mathbf{RGB} &= ( 329.70 \cdot (\text{temp} - 60)^{-0.1933},  \\
    &\quad  288.12 \cdot (\text{temp} - 60)^{-0.1155},  \\
    &\quad 255 )
\end{aligned}
\label{eq:awb_c}
\end{equation}

After computation, the RGB values are clipped to the range \([0, 255]\) to ensure valid color values. The resulting balanced RGB values are returned as a \texttt{float32} array, providing an accurate representation of the input temperature in RGB space.

\section{More Details of  Differential Camera Encoder}
\label{sec:network}

In the Differential Camera Encoder, an important aspect is the incorporation of the differences in camera setting scales. We extract the camera settings for \( F_r \) frames using the CLIP text encoder, compute the differences, and then reshape the result into an embedding of size \( F_r \times C \times H \times W \).

In addition, this section will also provide more details on the coarse embedding and the embedding encoder.

\subsection{Coarse Embedding}
The input to the coarse embedding is solely the provided camera settings. Based on a simplified version of the physical simulation model, it outputs an embedding with a shape of \( F_r \times C \times H \times W \).

For bokeh rendering, the input bokeh blur parameter is treated as an equivalent Gaussian blur kernel. A larger parameter indicates that the weight of each pixel in the output is lower, resulting in smaller global pixel embedding values.

\begin{figure}[h]
\centering
\includegraphics[width=0.99\linewidth, trim={0 0 0 0}, clip]{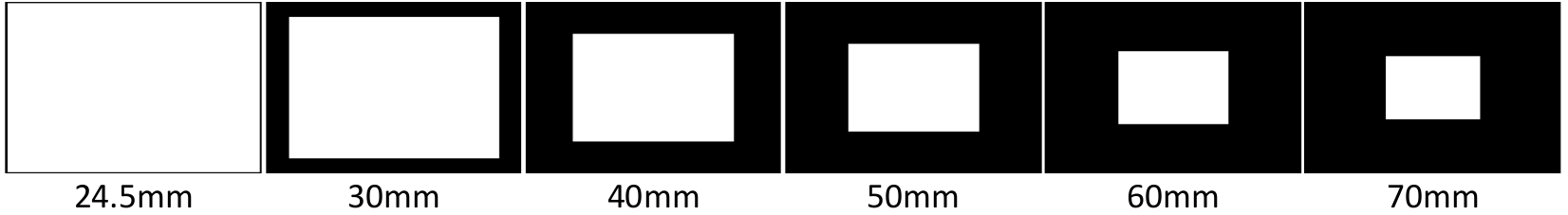}
\caption{\small{We use a mask as the coarse embedding for focal length control. The black areas represent pixels around the edges of the frame that should not be displayed at the given focal length.}}
\label{fig:zoom_mask}
\end{figure}

As illustrated in Fig. \ref{fig:zoom_mask}, for focal length, we use mask to proxy the coarse embedding. Specifically, after calculating the field of view (FoV) ratio, we mask out regions of the original image resolution that should not be present.

For shutter speed, we roughly estimate the ratio between the target shutter time and the base shutter time (simplified as 0.2 second on average). This ratio is then used to compute the overall average brightness ratio of the image, which serves as the global coefficient for the coarse embedding.

For color temperature, we estimate the ratio coefficients for the RGB channels based on the color temperature value, using a simplified version of the corresponding formula from Equation \ref{eq:awb_a} to Equation \ref{eq:awb_c}. These coefficients are then used as the scaling factors for the coarse embedding.

\subsection{Embedding Encoder}
The embedding encoder takes both the coarse embedding and the differential information embedding as input. After encoding, it injects the information into the temporal attention layers of the foundation model in a hierarchical manner. Its internal structure is based on the T2I adapter \cite{Mou_2023_T2Iadapter}, with additional temporal structures for multi-setting processing.

\section{More Details of Proposed Metrics}
\label{sec:metrics}

\subsection{Accuracy}
To evaluate the accuracy of the camera physics in generated images, we first simulate the reference frames of the base image under multiple camera settings, using the same scene description and corresponding camera parameters for generation. We then calculate the overall trend of camera effects within the reference frames and the overall trend of camera effects within the generated multi-frame sequence. The Pearson correlation coefficient between these two trends is computed as an accuracy metric (CorrCoef). For each type of camera setting, we employ different methods to calculate the camera effects.

\begin{itemize}
    \item For Bokeh: We compute the average blur level per frame using the Laplacian operator.
    \item For Focal Length: We first detect feature points using SIFT \cite{Lowe_2004_SIFT}, then perform feature matching between adjacent frames using Brute-Force Matcher \cite{Matcher}. We calculate the similarity transformation matrix from the matched points and extract the scaling factor from the transformation matrix.
    \item For Shutter Speed: We compute the average brightness per frame.
    \item For Color Temperature: We compute the average color per frame.
\end{itemize}

\subsection{Consistency}

For the consistency between frames corresponding to different camera setting values, we compute the frame-to-frame consistency using the Frame-wise Learned Perceptual Image Patch Similarity (LPIPS) \cite{Zhang_2018_Perceptual}. Subsequently, we average the LPIPS scores of all adjacent frames to obtain the final score. An important nuance here is that a lower LPIPS score is not always preferable, as we require some variation in camera effects. Therefore, the LPIPS score should be compared to that of reference videos, with a closer match indicating better performance.

\subsection{Following}
We measure the prompt following of the generated frames by evaluating their alignment with the input prompts. Specifically, we use the CLIP \cite{Radford_2021_CLIP} text and image encoders to obtain the features of the prompt and the generated frame, and then compute the cosine similarity between the two.

\section{More Visual Results}
\label{sec:results}
In this section, we provide additional visual results and comparisons with other methods.

\begin{figure*}[h]
\centering
\includegraphics[width=1\linewidth, trim={0 0 0 0}, clip]{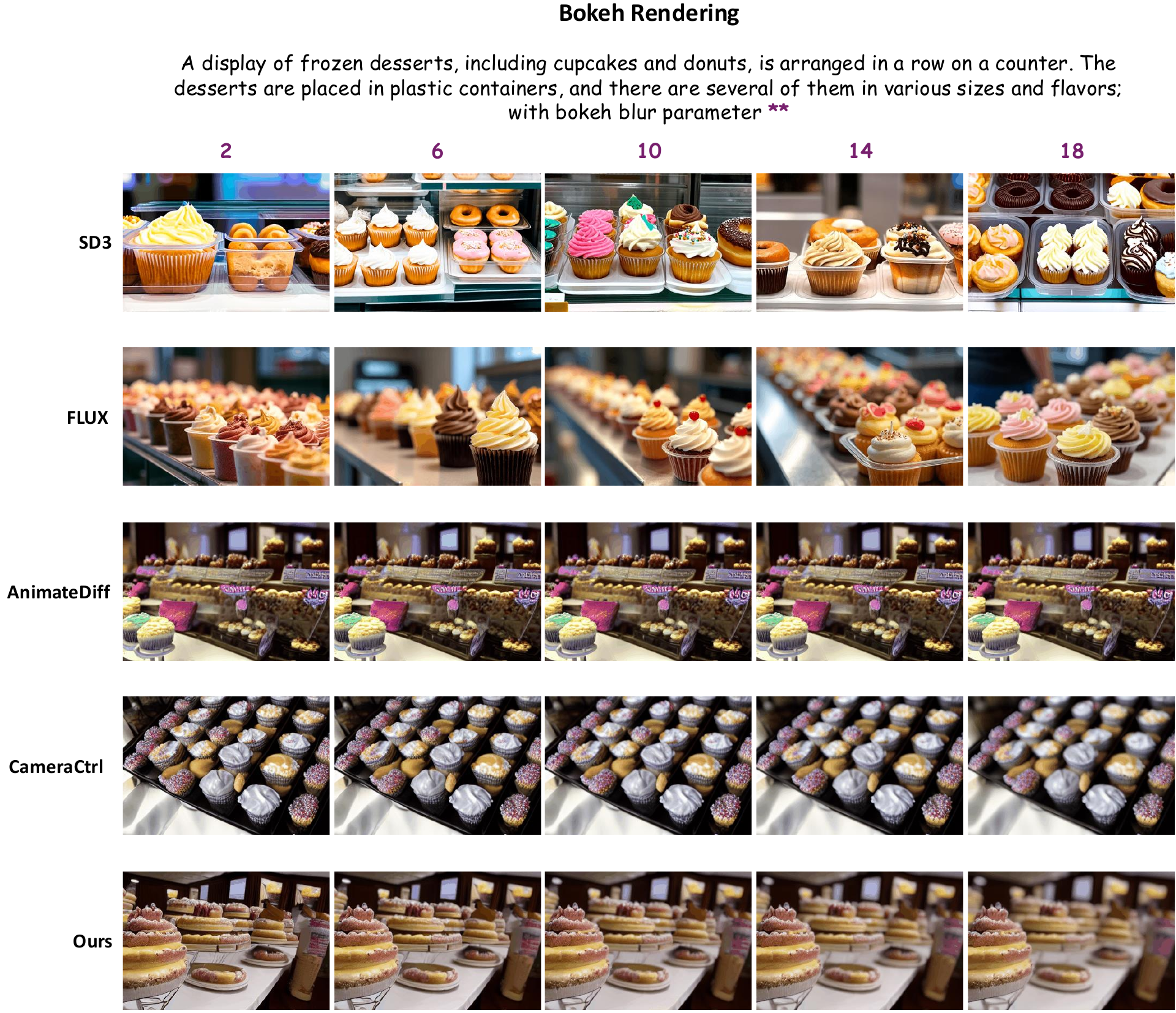}
\caption{\small{Visual comparisons between different generative methods on camera bokeh rendering control. Both AnimateDiff \cite{Guo_2023_AnimateDiff} and CameraCtrl \cite{He_2024_Cameractrl} have been fine-tuned/trained on our data.}}
\label{fig:result_bokeh}
\end{figure*}

\begin{figure*}[h]
\centering
\includegraphics[width=1\linewidth, trim={0 0 0 0}, clip]{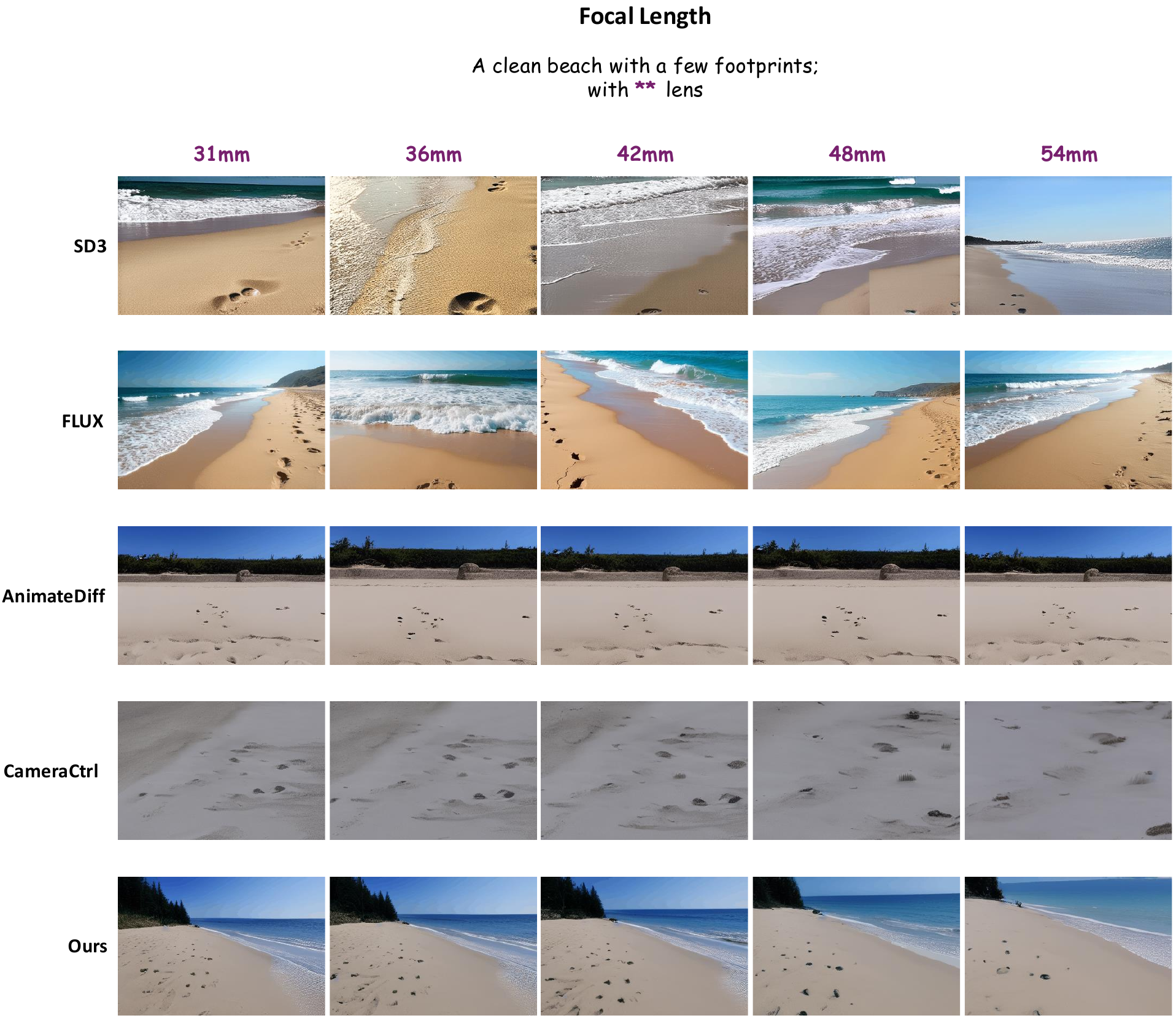}
\caption{\small{Visual comparisons between different generative methods on camera focal length control. Both AnimateDiff \cite{Guo_2023_AnimateDiff} and CameraCtrl \cite{He_2024_Cameractrl} have been fine-tuned/trained on our data.}}
\label{fig:result_focal_length}
\end{figure*}

\begin{figure*}[h]
\centering
\includegraphics[width=1\linewidth, trim={0 0 0 0}, clip]{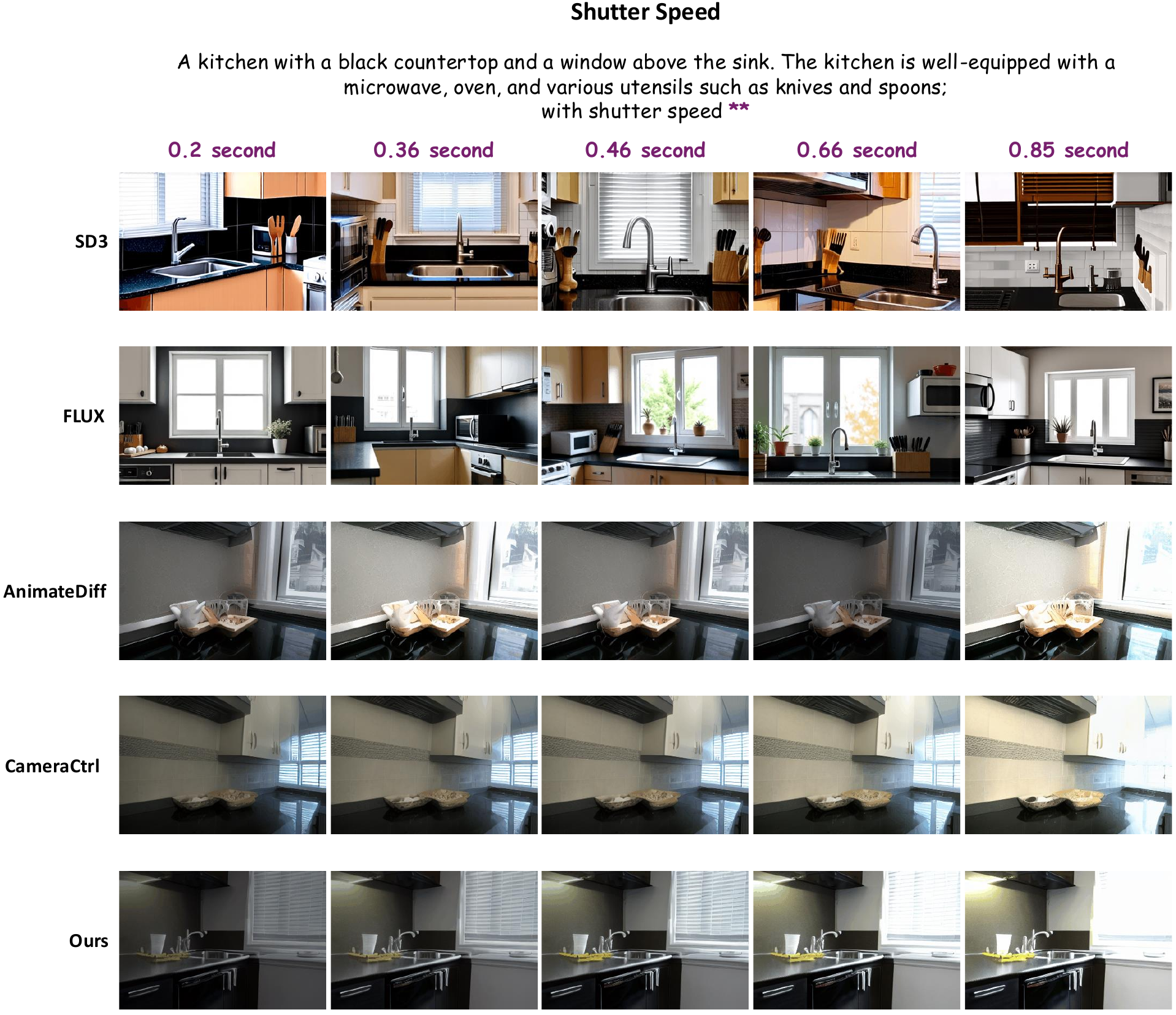}
\caption{\small{Visual comparisons between different generative methods on camera shutter speed control. Both AnimateDiff \cite{Guo_2023_AnimateDiff} and CameraCtrl \cite{He_2024_Cameractrl} have been fine-tuned/trained on our data.}}
\label{fig:result_shutter_speed}
\end{figure*}

\begin{figure*}[h]
\centering
\includegraphics[width=1\linewidth, trim={0 0 0 0}, clip]{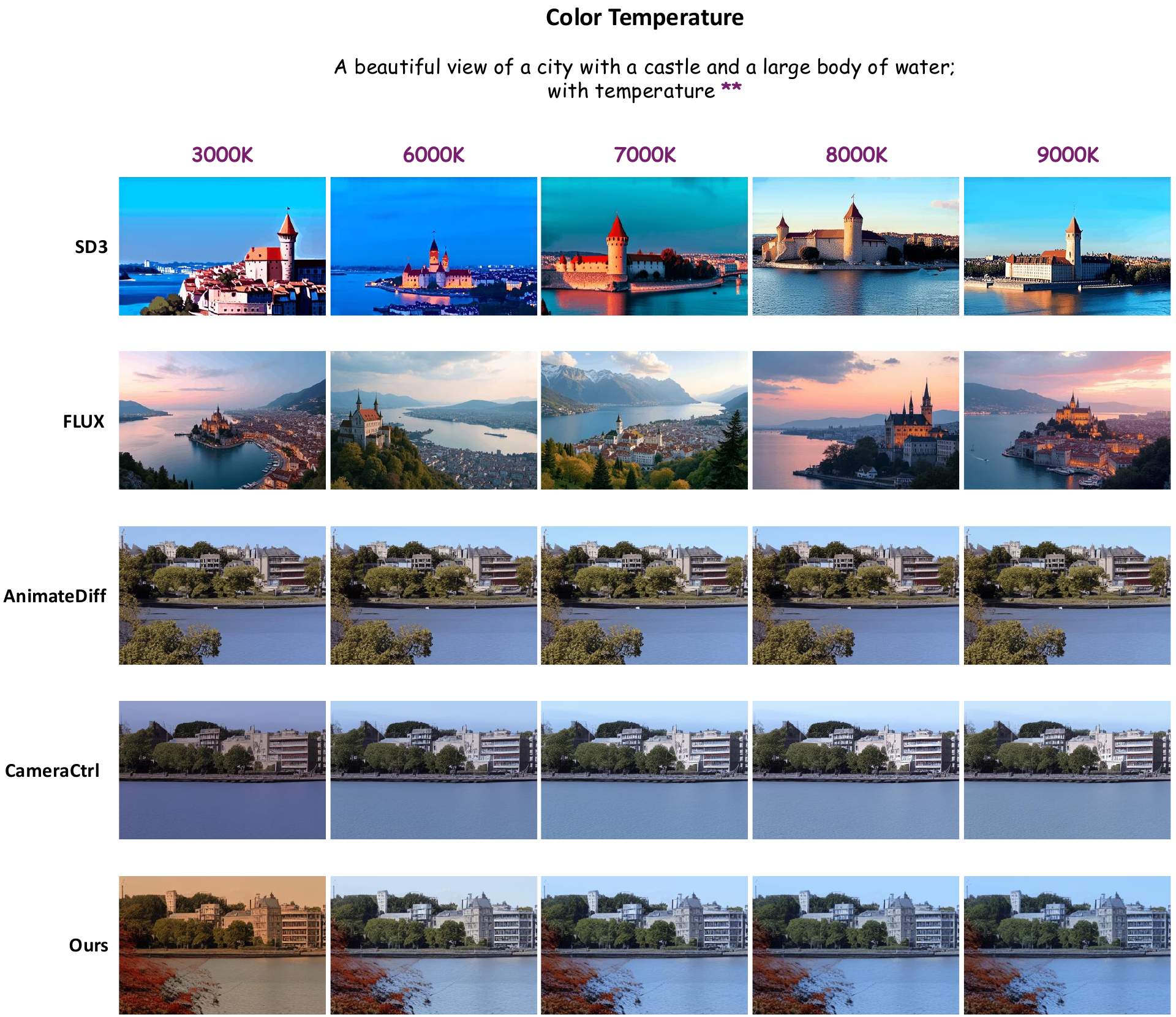}
\caption{\small{Visual comparisons between different generative methods on camera color temperature control. Both AnimateDiff \cite{Guo_2023_AnimateDiff} and CameraCtrl \cite{He_2024_Cameractrl} have been fine-tuned/trained on our data.}}
\label{fig:result_color_temperature}
\end{figure*}

Fig. \ref{fig:result_bokeh} to Fig. \ref{fig:result_color_temperature} illustrate the visual comparisons for bokeh rendering, focal length, shutter speed, and color temperature across various generative methods. Our approach demonstrates significant advantages in understanding camera physical parameters while maintaining scene consistency.

We strongly encourage readers to view the videos/GIFs we provide for more intuitive comparisons and additional case studies.

\end{document}